\documentclass[conference,a4paper]{IEEEtran}
\usepackage{color,amsmath,algorithm,algorithmic,graphicx,flushend}
\usepackage[usenames,dvipsnames,table]{xcolor}
\usepackage{fancyheadings}
\title{Autoencoding the Retrieval Relevance of Medical Images}
\author{Zehra \c{C}amlica$^1$,~H.R. Tizhoosh$^{1*}$,~Farzad Khalvati$^{2}$ \\
$^1$ KIMIA Lab, University of Waterloo, Canada, [*tizhoosh@uwaterloo.ca]\\ $^2$ Sunnybrook Research Institute, Toronto, Canada}

\begin{document}

\maketitle

\begin{abstract}
Content-based image retrieval (CBIR) of medical images is a crucial task that can contribute to a more reliable diagnosis if applied to big data. Recent advances in feature extraction and classification have enormously improved CBIR results for digital images. However, considering the increasing accessibility of big data in medical imaging, we are still in need of reducing both memory requirements and computational expenses of image retrieval systems.  This work proposes to exclude the features of image blocks that exhibit a low encoding error when learned by a $n/p/n$ autoencoder ($p\!<\!n$).  We examine the histogram of autoendcoding errors of image blocks for each image class to facilitate the decision which image regions, or roughly what percentage of an image perhaps, shall be declared relevant for the retrieval task. This leads to reduction of feature dimensionality and speeds up the retrieval process. To validate the proposed scheme, we employ local binary patterns (LBP) and support vector machines (SVM) which are both well-established approaches in CBIR research community. As well, we use IRMA dataset with 14,410 x-ray images as test data. The results show that the dimensionality of annotated feature vectors can be reduced by up to $50\%$ resulting in speedups greater than $27\%$ at expense of less than $1\%$ decrease in the accuracy of retrieval when validating the precision and recall of the top 20 hits.
\end{abstract}

\section{Problem and Motivation}
Searching for similar digital images in a given archive, or content-based image retrieval (CBIR), is necessary but a difficult task for several reasons. First of all, detecting similarity is a serious challenge. Answering the question what is similar to what is not easy when dealing with visual data at the pixel level. Secondly, measuring similarity is not an easy task either depending on how similarity has been quantified. And finally, searching in large archives takes time and can become infeasible when we are dealing with big data. 

Detecting and measuring similarities in large medical image archives may be a necessary task for diagnostic radiology, radiation oncology, cardiology and other clinical fields. Both the accuracy of retrieval and the speed of search become more significant in medical imaging as human life is in centre of attention. In contrast to non-medical images, the general appearance of a medical image may not be of interest where usually a certain part of the image, namely a region of interest (ROI) which could be an organ, a tumour or a specific tissue type is studied.  This implies that many (small) regions of the image may be irrelevant for a specific retrieval task. 

\section{The Idea}
The \textbf{idea} proposed in this paper is to detect irrelevant image blocks in each medical image class via analyzing the error histogram of a $n/p/n$ autoencoder in order to reduce the dimensionality of features for image retrieval.  We use an autoencoder with $p\!<\!n$ to ensure that autoencoding significant image blocks is accompanied with a large error, hence, making the detection of irrelevant blocks easier. Thus, the hypothesis of this paper is that the relevance of image blocks is directly proportional to the error of an autoencoder when the hidden layer is smaller than the input/output layer assuming that images are widely free from noise.

\section{Literature Review}
The literature on content-based image retrieval (CBIR) is vast and stretches over publications of more than 20 years. In following we briefly review major  CBIR works, elaborate on visual features for CBIR, and we also review some related works on autoencoders.
  
\textbf{CBIR --} Medical imaging devices are producing large number of images as they have become more sophisticated offering both higher acquisition speeds and resolution. Performing image search based on visual information, generally called content-based image retrieval (CBIR), has increasingly become more difficult in recent years. This is, on one hand, because of diverse challenges preventing accurate similarity detection. But on the other hand,  efficient processing of big data, or more precisely timely analysis of big image data, on ordinary computing devices with conventional algorithms appears to be a very daunting task. In the beginning era of digital image search, various searching methods were investigated, although researchers were mainly focused on text-based search to retrieve images \cite{Enser1995} \cite{Joshi2014}. Surveys provide an overview of literature \cite{Eakins2000}. The example of a complete overview of the first decade of research in this field is provided in \cite{Venters2000}. Also, medical image retrieval systems, as a special sub-field of CBIR, have been reviewed in \cite{Akgul2011}. CBIR systems help to retrieve, manage and navigate through huge visual data archives searchable when textual/visual queries are provided. Although CBIR systems differ in the methods applied to image in order to retrieve and store features and measure similarity, basic architectures of the systems are quite similar. Feature extraction and indexing or similarity measurement are two main processes of most CBIR systems.

 \textbf{Feature extraction --} The main purpose for extracting features is to create high-level descriptions from low-level data (pixel values). Recent medical image retrieval systems rely on visual features, such as color, shape, texture, and other spatial characteristics. Visual features are arranged in three levels: low level features (primitive), middle level features (logical) and high level features (abstract). Almost all early systems were based on low level features that capture characteristics (color, shape). But currently both mid-level(e.g. sub-image, bagging approach) and high-level (semantics) image representations are in demand. General visual features are implemented in most CBIR systems because of their independence from prior information and efficiency in computation \cite{132}. The efficiency of a CBIR system depends, among others,  on the quality of extracted features. If the features do not represent the image content adequately, similar images can hardly be retrieved. 
 
\textit{Gray level features --} Color is one of the most commonly used features in CBIR  \cite{Muller2004}\cite{54}. Considering local gray level features of each pixel in the image, global gray level features of the whole image can be formed. Building a gray-level histogram is a popular method to extract global  features from medical images \cite{54}\cite{32} \cite{57}. A typical histogram is discretized into 256 bins. Being independent from changes in resolution and rotations are special advantages of the histogram method. Additionally, being simple to implement, efficient to compute and having low space requirements are some reasons why histograms are common in CBIR \cite{15}\cite{47}. Nevertheless, possible assignments for similar color intensities to different bins \cite{32}  and the absence of any spatial information \cite{47} are main disadvantages of using histograms. To overcome these problems, partition-based histograms that contain spatial information by splitting the image into multiple partitions and calculating local histograms have also been developed \cite{31}. Moreover, to solve the spatial information problem in a histogram, the color coherence vectors (CCV) method has been proposed \cite{47}. It investigates similar gray level regions in the image and count the number of pixels of these regions. The method compares the number of pixels  in the region with a threshold and classifies them as coherent or incoherent. In this method, some spatial information is still missed. As well, determination of a threshold poses a potential problem.   Another method is the gray level correlogram which is supposed to extract both gray-level and spatial information from an image \cite{26}. The method processes pixel position, intensity, probability of intensity and distance.

\textit{Textural features --}  In medical imaging, textural features are one of most essential image features since gray levels may be incapable of effective object discrimination \cite{151}. As well, texture features may generally contain crucial information to diagnose  a disease. Smoothness, directionality, and randomness are some textural properties \cite{27}. There are different types of textural feature extraction methods, which are usually of  statistical, geometrical or model-based nature. Texture features can provide the means to classify \cite{146} and retrieve images \cite{147}. Energy, entropy, contrast and homogeneity are some typical values to characterize a texture in an image \cite{148}.  Statistical methods represent textures by the statistical distribution of the image intensity, such as co-occurrence matrices, Tamura features \cite{23}, Markov random field, fractal model, and multi-resolution filtering techniques. Furthermore, Local Binary Patterns (LBPs) have been implemented originally to describe texture of the images \cite{ldp-lbp-16}\cite{17}. LBP is a practical method to quantify the gray level textures by utilizing patterns of local neighborhoods. The LBPs  have been used in various applications for texture classification \cite{ldp-lbp-16} \cite{18}, face recognition \cite{20}, fingerprint identification \cite{12}, and automated cell phenotype image classification \cite{2}. In \cite{6} LBPs are used to characterize medical images, for instance magnetic resonances and mammography images. LBP is widely considered as the state-of-the-art texture descriptor because of low computational complexity and its invariance to changes in resolution. Recently, Tizhoosh introduced the concept of ``barcodes'' for image annotation that by using Radon transform may be a new binary approach to texture description \cite{Tizhoosh2015}.

\textbf{Multimodal searching --} After the feature extraction stage, each visual feature set is  usually stored in a vector.  Different strategies have been developed to use various modalities in searching for CBIR, such as constrained hierarchies or classes, early fusion and late fusion {\cite{14}\cite{59}. While the searching process is restricted with some hierarchies or classes in constrained methods, all images are searched in both early and late fusion methods. Using constrained method speeds up the retrieval process. Performing search within a local area (a certain class) or based on a hierarchical order takes less time than searching in the entire dataset. This method provides advantages especially for huge datasets as long as the class and hierarchy estimate do not fail \cite{local}. Image annotation and classification can be considered as a first step for speeding-up image retrieval in large databases. There are various approaches for image classification. SVM is a popular algorithm to perform reliable and generally fast classification \cite{unay1}\cite{Mueller2010}. For example, recently it has been proposed to use linear SVM methods with quadratic optimization method for CT brain images \cite{CT}. Also, SVM has been combined with K-NN classifiers \cite{Bishop} and with boosting \cite{boost}. Since we usually have to deal with high-dimensional feature space, most indexing methods cannot perform within reasonable time. Moreover, storage of these features constitutes another challenge for CBIR. Reduction of feature space without losing useful information is therefore a  crucial step for both image annotation and retrieval. 
   
\textbf{Autoencoders --} 
Autoencoders are a a special type of neural networks to decode the encode inputs with minimum error. Introduced by Hinton et al. \cite{Rumelhart1986} to make backpropagation networks work without a teacher, autoencoders provide a very sophisticated unsupervised learning scheme. For instance,  the denoising autoencoder can be trained to reconstruct a data from one of its corrupted versions \cite{Vincent2008}. Very deep autoencoders can be initialized by learning many layers of features on color images \cite{Krizhevsky2011}. Autoencoders can then map images to short binary codes. As well, auoendocoders have been applied to compress mammograms by using image patches instead of the entire image \cite{Tan2011}. \cite{Valentine2012} adapt the autoencoder to the continuous case and use  autoencoders for seismic waveforms, and offer a demonstration in which they compress 512-point waveforms to 32-element encodings. \cite{Bengio2013} put the use of deep learning and autoencoders in perspective by a detailed and comprehensive investigation of representation learning. \cite{Wang2014} attempts to propose a generalized autoencoder via manifold learning and uses it for digit and face image manifolds. \cite{Baldi2012} presents a general mathematical framework for the study of both linear and non-linear autoencoders. \cite{Lu2014} propose a feature ensemble learning method based on sparse autoencoders for image classification. To our knowledge, no work has proposed to use the autoencoding error to quantify the retrieval relevance of images in general and medical images in particular. As described in the next section, we train a \emph{shallow} autoencoder and record the error histogram of each class to eliminate image blocks that are irrelevant for retrieval.

\section{Our Approach: Autoencoding Relevance}
Not every pixel and not every region of an image may be relevant for a specific retrieval task. This becomes particularly significant for medical imaging where most of the time a certain region of interest, ROI, needs to be analyzed (e.g. a tumour, an organ). As a result, we envision an approach that attempts to eliminate small patches (blocks) of the image from feature extraction process. This selective reduction of image patches must be based on some universal criterion of relevance to ensure we have a generic approach that can be trained for different image modalities and specific ROIs. 

Autoencoders, with $n/p/n$ architecture, encode $n$ inputs into $p$ positions, and then decode $p$ positions back into $n$ outputs. Generally, we may use $p\!<\!n$ in which case autoencoder functions as a compressor to reduce dimensionality. Such an autoencoder is basically a \emph{shallow} neural network with some level of error. The error can be reduced if we \emph{deepen} the network, for instance make it an $n/m/k/p/k/m/n$ autoencoder with $n\!<\!m\!<\!k<\!p$. However, this is not what we intend to do. We would like to design a shallow network to keep the decoding error high. But why?

Using a shallow network, a $n/p/n$ autoencoder, with a relatively high error for decoding image blocks, captured in a histogram matrix $\mathbf{H}$ for each class, will enable us to locate retrieval-irrelevant image regions. If the image block contains complex structures (edges, textures etc.), then the encoding error is expected to be high for $n/p/n$ autoencoder specially when we ascertain that $p\!<\!n$. In contrast, if we are encoding blocks of uniform regions with no significant gradient change, then we shall expect low decoding error. The blocks, therefore, with lowest decoding errors are the ones with the least contribution to accurate retrieval (or so is our assumption to be validated in experimentation).  

If we divide image $I$ into $k\times k$ blocks and desire that $d\in [0,1)$ fraction of the image area be reduced for retrieval-oriented feature extraction, then the task is to eliminate as many as $\lfloor d\times k\times k\rfloor$ blocks by not extracting features from them. If we record the autoencoding errors for all blocks of a certain image class, then this can be done by ignoring the blocks below an error threshold that eliminates $\lfloor d\times k\times k\rfloor$ blocks. Of course, this assumes that an inter-class method, e.g. SVM, has already classified the query image and assigned it to a certain class. The proposed reduction of feature dimensionality using autoencoding error analysis occurs to improve the intra-class retrieval task.  Figure \ref{fig:blockauto} illustrates the idea of relevance quantification via  autoencoding error histogram. As well, Algorithm \ref{alg:approach} describes the proposed approach. In order to implement a complete solution, we use LBP features and SVM to classify the images. One can, in future works, investigate the use of \emph{opposites} as already reported in iterative for learning and optimization in order to see the effect of using a network incorporating opposites \cite{Tizhoosh2007a,Tizhoosh2007b,Tizhoosh2008}. 

\begin{figure*}[htb]
\center
\includegraphics[width=0.9\textwidth]{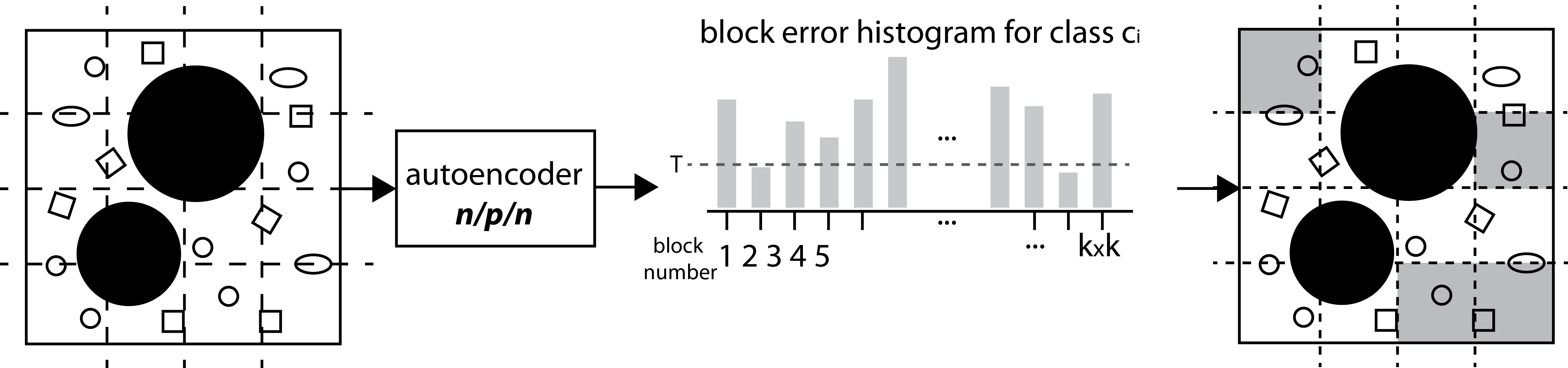}
\caption{Schematic illustration of the proposed approach: Image blocks are autoencoded in an $n/p/n$ architecture. The encoding error of each block is recorded for each image class to create the error histogram. A desired reduction (e.g. $25\%=4$ blocks) can be used to establish a threshold $T$ in order to exclude a number of blocks (gray blocks) from feature extraction. }
\label{fig:blockauto}
\end{figure*}

\begin{algorithm}[tb]
\caption{Proposed approach}
\begin{algorithmic}[1]
\label{alg:approach}
\STATE ------- \textbf{Configuration} -------
\STATE Set $k$ to divide the image into $k\times k$ blocks 
\STATE Set the desired reduction rate $d \in [0,1)$
\STATE Set $n/p/n$ for the autoencoder ($p<n$)
\STATE ------- \textbf{Training} -------
\STATE Get the number of training images $m_{\textrm{max}}$
\STATE Initialize the feature matrix $\mathbf{F}$ 
\STATE Initialize the class vector $\mathbf{c}$
\STATE Initialize the error histogram $\mathbf{H}$ for all classes
\FOR{\textbf{each} $i\in \{1,2,\dots,m_{\textrm{max}}\}$ }
	\STATE Read the training image $\mathbf{I}_i$ and its class $c_i$
	\STATE  $\mathbf{c}\leftarrow c_i$
	\FOR{\textbf{each} $j\in \{1,2,\dots,k\times k\}$ }
		\STATE $\mathbf{B}_j \leftarrow$ \emph{currentBlock}$(\mathbf{I}_i)$ 
				\STATE $\mathbf{f}\leftarrow$ \emph{extractLBPfeatures}$(\mathbf{B}_j)$
		\STATE  $\mathbf{F}\leftarrow$ \emph{appendFeatures}$(\mathbf{F},\mathbf{f})$
		\STATE \emph{error}$ \leftarrow$\emph{autoEncode}$(\mathbf{B}_j)$
				\STATE $\mathbf{H}(c_i,j) \leftarrow$ $\mathbf{H}(c_i,j) +$ \emph{error}
	\ENDFOR 
\ENDFOR
\STATE $[\mathbf{v}_1,\mathbf{v}_2,\dots] \leftarrow$ \emph{TrainSVM}$(\mathbf{F},\mathbf{c})$
\STATE Save support vectors $\mathbf{v}_1,\mathbf{v}_2,\dots$
\STATE Save the error histogram $\mathbf{H}$
\STATE ------- \textbf{Testing} -------
\STATE Read $\mathbf{v}_1,\mathbf{v}_2,\dots$ and $\mathbf{H}$ and a new image $\mathbf{I}_\textrm{new}$ 
\FOR{\textbf{each} $j\in \{1,2,\dots,k\times k\}$ }
	\STATE $\mathbf{B}_j \leftarrow$ \emph{currentBlock}$(\mathbf{I}_\textrm{new})$ 
	\STATE $\mathbf{f}\leftarrow$ \emph{extractLBPfeatures}$(\mathbf{B}_j)$
\ENDFOR 
\STATE $c_\textrm{new} \leftarrow $ \emph{classifySVM} $(\mathbf{f})$
\STATE $\mathbf{f'}\leftarrow$ \emph{ignoreBlocks}$(\mathbf{f},\mathbf{H},c_\textrm{new})$
\STATE $<I_1^*, I_2^*, I_3^*, \cdots> \leftarrow$ \emph{calculateSimilarity}$(\mathbf{F},\mathbf{f'},\mathbf{H},c_\textrm{new})$
\STATE Show retrieved images $<I_1^*, I_2^*, I_3^*, \cdots> $
\end{algorithmic}
\end{algorithm}

\section{Experiments and Results}
In this section, we first provide information about the benchmark data. The error measurement for classification is described next. Subsequently, we detail the accuracy measures for the retrieval task. The settings for LBP and SVM are described afterward. Lastly, the results will be reported. 

\textbf{Image Dataset --} The Image Retrieval in Medical Applications (IRMA) 2009 database is a collection of 14,410 x-ray images that have been randomly collected from daily routine work at the Department of Diagnostic Radiology of the RWTH Aachen University (Fig. \ref{fig:samples}).  The downscaled images were collected from different ages, genders, view positions, and pathologies~\cite{Mueller2010}. Each image in the dataset has an IRMA code. According to these codes, 193 classes are defined according to 2008 IRMA codes. The IRMA code comprises four axes with three to four positions each: 1) the technical code (T) (modality), 2) the directional code (D) (body orientations), 3) the anatomical code (A) (body region), and 4) the biological code (B) (the biological system examined). The complete IRMA code consists of 13 characters TTTT-DDD-AAA-BBB, with each character in $\{0,\dots,9; a,\dots,z\}$. As many as 12,677 images are separated for training. The remaining 1,733 images are used as test data. In this project, the IRMA 2009 dataset has been used with specified 2008 IRMA labels (consisting of 193 classes) for retrieval purposes. Otherwise, same dataset is utilized with general 2005 IRMA labels (consisting of 57 classes) for classification purpose. 2005 IRMA labels are more general than 2008 IRMA labels because it has been made of 6 characters from top of hierarchical classes, TT-D-AA-B. In 2009 dataset, each image can not have been coded according to 2005 IRMA coding regularity. A total number of 12,631 images from training set and 1,639 images from testing set have 2005 IRMA codes. For this reason, SVM classification is implemented on corresponding images.
\begin{figure}[tb]
\begin{center}
\includegraphics[width=0.98in,height=0.98in]{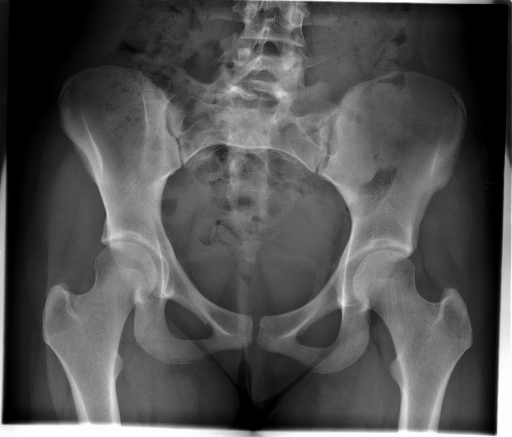}
\includegraphics[width=0.98in,height=0.98in]{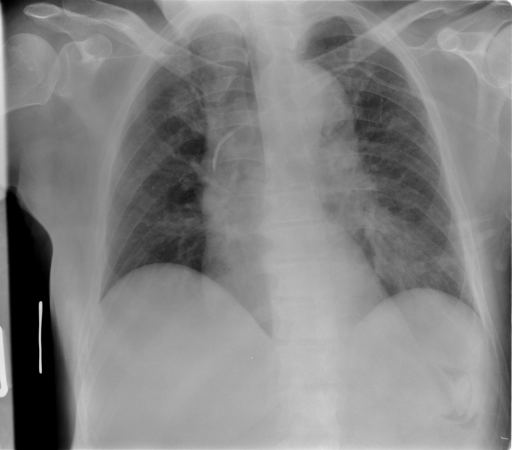}
\includegraphics[width=0.98in,height=0.98in]{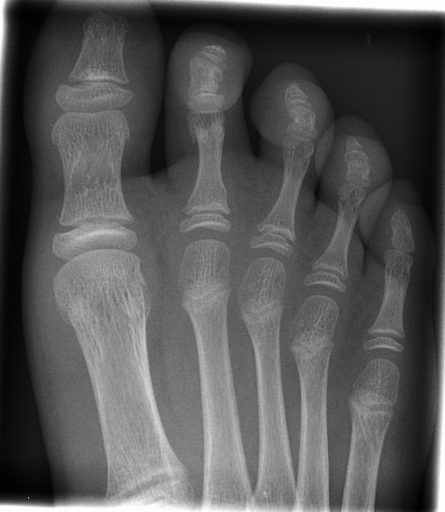}
\includegraphics[width=0.98in,height=0.98in]{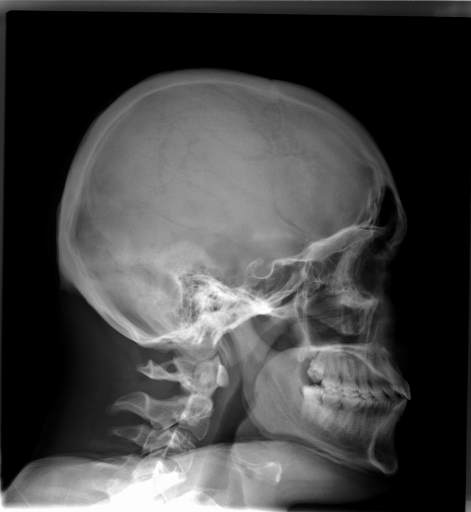}
\includegraphics[width=0.98in,height=0.98in]{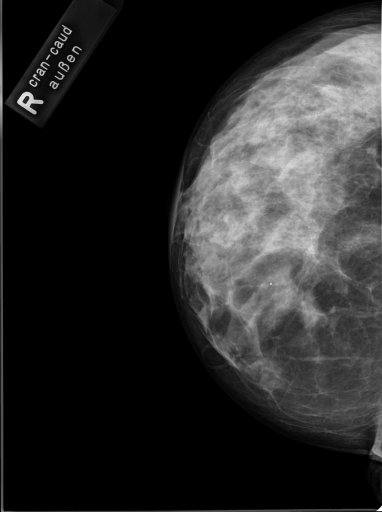}
\includegraphics[width=0.98in,height=0.98in]{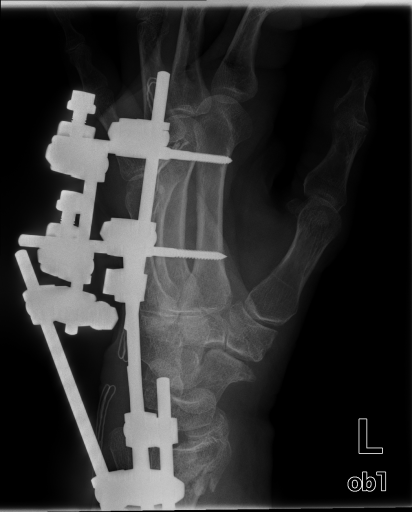}
\caption{Sample images from IRMA dataset.}
\label{fig:samples}
\end{center}
\end{figure}

\textbf{Error Measurement for Classification --} The ImageCLEF project has defined an error score evaluation method in order to evaluate the classification performance of methods on IRMA dataset \cite{Mueller2010}. As in IRMA dataset all images are labelled with the technical, directional, anatomical and biological independent axes, the error $E$ can be defined as follows
\begin{equation}
E = \sum\limits_{i=1}^n\dfrac{1}{b_{i}}\dfrac{1}{i}\delta(I_{i},\hat{I_{i}})
\label{6}
\end{equation}
 where $b_{i}$ is number of possible labels at position $i$ and $\delta$ is the decision function delivering $1$ for wrong label and $0$ for correct label when the IRMA codes of the image $I_i$ is compared with the IRMA code of the image $\hat{I}_i$. For every axis, the maximal possible error is computed and the errors are normalized between $0.25$ and $0$. If all positions in all axes are wrong, error value is $1$.

\textbf{Accuracy Measurement for Retrieval --} Looking at the top $m$ retrieved images, the number of correctly retrieved images (true positives $T_p$) and wrongly retrieved images (false positives $F_p$) can be used to calculate the precision $P_\textrm{top m}$ of the retrieval: $P_\textrm{top m} = \frac{T_p}{T_p+F_p}$. Analogously, using the top $m$ retrieved images, the number of correctly retrieved images (true positives $T_p$) and wrongly not-retrieved images (false negative $F_n$) can be used to calculate the recall $R_\textrm{top m}$ of the retrieval: $ R_\textrm{top m} = \frac{T_p}{T_p + F_n}$.

\textbf{LBP and SVM Settings --} We extracted local binary patterns from $3\times 3$ neighbourhoods within each image block, converted the binary numbers to decimal numbers to calculate a histogram $\mathbf{h}_{\textrm{LBP}}$. These histograms were used for both classification and retrieval. To classify the input image we used support vector machines (SVM). The LBP histogram features form IRMA training dataset are used to train the multi-class SVM with radial basis function as its kernel \cite{libsvm}. 

\textbf{Results --} We first classified the images with SVM using LBP features. We tested to extract LBP features for the entire image but that led to a significant decrease in classification accuracy ($\approx 50\%$). Analog to global versus local thresholding, it appears that calculating LBP histograms for image blocks is more capable of capturing the spatial characteristics of the image compared to extracting only one LBP histogram for the entire image. As Table \ref{table:allresults1} illustrates, the LBP-SVM approach achieves the lowest error score and hence the highest accuracy for $4\times 4$ blocks (image divided into 16 regions). 

\begin{table}[htdp]
\caption{SVM accuracy with LBP features for classification of 2005 IRMA image dataset  containing 14,270 images constituting 57 image categories.}
\begin{center}
\begin{tabular}{|c|c|c|}
Blocks & Accuracy & Error score \\ \hline
$4\times 4$ 	& $75.50\%$ & 116.77 \\
$5\times 5$ 	& $74.31\%$ & 121.34 \\
$6\times 6$ 	& $72.23\%$ & 133.90 \\ \hline
\end{tabular}
\end{center}
\label{table:allresults1}
\end{table}%

For the retrieval, we autoencoded the image blocks first using the restricted Boltzmann machine (RBM) function. We did try different values for $p$ but as long as $p\!<\!n$ was maintained, the error levels were not considerably affected. The number of iterations for the autoencoder was set to 5 as we observed that more iterations did not change our results. We used cross correlation to measure the similarity of two feature vectors for two images, and we compared the cases with no reduction (all mage blocks considered for feature calculation) and different reduction levels (1/8, 1/4 and 1/2 corresponding to 12.5\%, 25\% and 50\%, respectively). The precision and recall were calculated when the top 10, 20 and 30 images were in focus. Retrieval times were recorded as well. Table \ref{table:allresults2} provides the averages of 100 runs for different settings. Table \ref{table:allresults3} summarizes the results. It is obvious that for finer grid structures ($6\times 6$ blocks) the time savings of greater than $27\%$ can be achieved where $50\%$ of the image blocks have been ignored resulting in $50\%$ reduction of the feature vector size. This becomes a significant result when we observe a slight decrease in accuracy (both precision and recall) less than $1\%$ for the top 20 hits of the retrieval. One should note that a full, one-to-one translation of the space savings (namely $50\%$) into computational savings (here $27\%$) may not be possible because of the intrinsic difference in space-time  relationship  and with respect to specifications of the algorithmic steps involved in saving and processing tasks.

\begin{table*}[htdp]
\caption{The average precision $P$, recall $R$ and time $t$ are measured for top 10, top 20 and top 30 retrieval for randomly selected image from 57 classes and repeated 100 times. As distance metric, cross-correlation was used.}
\begin{center}
\begin{tabular}{|c|c||c|c||c|c||c|c||c|}
Blocks & Reduction &  $P_\textrm{Top 10}$   & $R_\textrm{Top 10}$ & $P_\textrm{Top 20}$  & $R_\textrm{Top 20}$ & $P_\textrm{Top 30}$  & $R_\textrm{Top 30}$ &  $t$(sec) \\ \hline
$4\times 4$  & 0 	& 0.872 & 0.1807 & 0.868 & 0.2527 & 0.882 & 0.3214 & 0.01807 \\
$5\times 5$ & 0 	& 0.874 & 0.1783 & 0.873 & 0.2538 & 0.883 & 0.3222 & 0.02154 \\
$6\times 6$ & 0 	& 0.871 & 0.1792 & 0.877 & 0.2542 & 0.882 & 0.3220 & 0.02620 \\ \hline
$4\times 4$  & 1/8 	& 0.862 & 0.1761 & 0.866 & 0.2485 & 0.878 & 0.3199 & 0.01684 \\
 $5\times 5$ & $\approx$1/8 & 0.869 & 0.1783 & 0.872 & 0.2531 & 0.880 & 0.3212 & 0.02005 \\
$6\times 6$ & $\approx$1/8 & 0.867 & 0.1786 & 0.873 & 0.2519 & 0.882 & 0.3221 & 0.02388 \\ \hline
$4\times 4$  & 1/4 	& 0.863 & 0.1762 & 0.862 & 0.2478 & 0.877 & 0.3205 & 0.01592 \\
$5\times 5$ & $\approx$1/4 & 0.870 & 0.1788 & 0.870 & 0.2497 & 0.879 & 0.3214 & 0.01833 \\
$6\times 6$ & 1/4 	& 0.869 & 0.1775 & 0.871 & 0.2513 & 0.879 & 0.3195 & 0.02182 \\ \hline
$4\times 4$  & 1/2 	& 0.856 & 0.1745 & 0.858 & 0.2458 & 0.865 & 0.3147 & 0.01396 \\
$5\times 5$ & $\approx$1/2 & 0.864 & 0.1762 & 0.862 & 0.2491 & 0.875 & 0.3185 & 0.01591 \\
$6\times 6$ & 1/2 	& 0.868 & 0.1790 & 0.870 & 0.2519 & 0.878 & 0.3212 & 0.01905 \\ \hline
\end{tabular}
\end{center}
\label{table:allresults2}
\end{table*}%

\begin{table*}[htdp]
\caption{Decrease in precision/recall and gain in time (for top 20 hits) compared to no reduction case where the LBP is used for retrieval comparisons.}
\begin{center}
\begin{tabular}{|c|c|c|c|c|c|c|c|c|c|}
			& \multicolumn{3}{c|}{$12.5\%$ reduction}  &  \multicolumn{3}{c|}{$25\%$ reduction}  & \multicolumn{3}{c|}{$50\%$ reduction}  \\
Blocks  & Precision  & Recall  & Time  & Precision  & Recall  & Time  & Precision  & Recall  & Time \\ \hline
 $4\times 4$  & 0.20\%  & 1.6\%  & 6.8\%  & 0.7\%  & 1.9\%  & 11.8\%  & 1.2\%  & 2.7\%  & 22.70\% \\
 $5\times 5$  & \cellcolor[gray]{.8}0.11\%  &  \cellcolor[gray]{.8}0.2\%  &  \cellcolor[gray]{.8}6.9\%  & 0.3\%  & 1.2\%  & 14.9\%  & 1.3\%  & 1.8\%  & 26.13\% \\
 $6\times 6$  & 0.40\%  & 0.2\%  & 8.8\%  & 0.7\%  & 1.1\%  & 16.7\%  &  \cellcolor[gray]{.8}0.8\%  &  \cellcolor[gray]{.8}0.9\%  &  \cellcolor[gray]{.8}27.20\% \\ \hline
\end{tabular}
\end{center}
\label{table:allresults3}
\end{table*}%

\section{Summary}
Searching for similar images in large medical image archives is both necessary and challenging. Whereas we can classify the query image in a very short time to assign it to an existing image category, the actual retrieval of similar images may need more computational resources through more costly and one-by-one comparisons. This becomes a serious obstacle for medical imaging with emerging big image data availability. There are many methods to reduce the dimensionality of image classification and retrieval tasks. Autoencoders have been investigated in the past with respect to their compression capabilities. In this work, we proposed a different approach to data reduction. Motivated by the fact that in medical image analysis usually a certain region of interest, ROI, is in focus of user evaluation, we proposed to eliminate some image patches (rectangular blocks) from feature extraction process. This leads to reduction of both memory requirements and computational expense of the retrieval task. To decide which image blocks are rather irrelevant for the retrieval process, we trained a $n/p/n$ autoencoder ($p\!<\!n$) with the image blocks as both in- and output. We recorded the autoencoding errors in a histogram for each image class.  This histogram is then thresholded to exclude a certain percentage of the image area (which has low autoencoding error and does not contribute to image retrieval task), in terms of number of image blocks, for each new image. Experiments with IRMA dataset with 14,410 x-ray images showed that, accepting a slight decrease in precision and recall for the top 20 hits, the space requirements for the annotated feature vectors can be cut down by $50\%$ where simultaneously the speed of the retrieval can be increased by $27\%$.

\bibliographystyle{IEEEbib}
\bibliography{refs}

\begin{thebibliography}{10}

\bibitem{Enser1995}
P.G.B. Enser,
\newblock ``Pictorial information retrieval,''
\newblock {\em J. Document}, vol. 51, no. 2, pp. 126--170, 1995.

\bibitem{Joshi2014}
MD~Joshi, R.M. Deshmukh, K.N. Hemke, A.~Bhake, and R.~Wajgi,
\newblock ``Image retrieval and re-ranking techniques - a survey,''
\newblock {\em Signal and Image Processing: An International Journal (SIPIJ)},
  vol. 5, no. 2, 2014.

\bibitem{Eakins2000}
J.P. Eakins and M.E. Graham,
\newblock {\em Content-based image retrieval, Tech. Rep. JTAP-039},
\newblock ISC Technology Application Program, 2000.

\bibitem{Venters2000}
C.C. Venters and M.~Cooper,
\newblock {\em Content-based image retrieval, Tech. Rep. JTAP-054},
\newblock JISC Technology Application Program, 2000.

\bibitem{Akgul2011}
Akgul C.B., Rubin D., Napel S., Beaulieu C., Greenspan H., and Acar B.,
\newblock ``Content-based image retrieval in radiology: Current status and
  future directions,''
\newblock {\em J. Digital Imaging}, vol. 24 (2), pp. 208--222, 2011.

\bibitem{132}
Y.K. Chan and C.Y. Chen,
\newblock ``Image retrieval system based on color complexity and color spatial
  features,''
\newblock {\em J. of Systems and Software}, vol. 71, no. 1, pp. 65--70, 2004.

\bibitem{Muller2004}
H.~M{\"u}ller, N.~Michoux, and D.~Bandon,
\newblock ``A review of content-based image retrieval systems in medical
  applications�--clinical benefits and future directions,''
\newblock {\em Int. J. of Medical Informatics}, vol. 73, no. 1, pp. 1 -- 23,
  2004.

\bibitem{54}
D.C. Shi, L.~Xu, and L.Y. Han,
\newblock ``Image retrieval using both color and texture features,''
\newblock {\em The Journal of China Universities of Posts and
  Telecommunications}, vol. 14, pp. 94 -- 99, 2007.

\bibitem{32}
Konstantinidis K., Gasteratos A., and Andreadis I.,
\newblock ``Image retrieval based on fuzzy color histogram processing,''
\newblock {\em Optics Communications}, vol. 248, no. 4–6, pp. 375 -- 386,
  2005.

\bibitem{57}
J.R. Smith and S.F. Chang,
\newblock ``Tools and techniques for color image retrieval,''
\newblock in {\em IS\&T/SPIE Proceedings}, 1996, pp. 426--437.

\bibitem{15}
G.~Gagaudakis and P.L. Rosin,
\newblock ``Incorporating shape into histograms for {CBIR},''
\newblock {\em Pattern Recognition}, vol. 35, no. 1, pp. 81--91, 2002.

\bibitem{47}
G.~Pass, R.~Zabih, and J.~Miller,
\newblock ``Comparing images using color coherence vectors,''
\newblock in {\em The Fourth ACM International Conference on Multimedia}, 1996,
  pp. 65--73.

\bibitem{31}
Tan K.L., Ooi B.C., and Yee C.Y.,
\newblock ``An evaluation of color-spatial retrieval techniques for large image
  databases,''
\newblock {\em Multimedia Tools Appl.}, vol. 14, no. 1, pp. 55--78, 2001.

\bibitem{26}
J.~Huang, S.R. Kumar, M.~Mitra, W.J. Zhu, and R.~Zabih,
\newblock ``Image indexing using color correlograms,''
\newblock in {\em Proceedings of the 1997 Conference on Computer Vision and
  Pattern Recognition (CVPR '97)}, 1997, CVPR '97, pp. 762--.

\bibitem{151}
S.~Radhouani, J.~H. Lim, J.-P. Chevallet, and G.~Falquet,
\newblock ``Combining textual and visual ontologies to solve medical multimodal
  queries,''
\newblock in {\em IEEE International Conference on Multimedia and Expo}, 2006,
  pp. 1853--1856.

\bibitem{27}
Po-Whei Huang and S.~K. Dai,
\newblock ``Image retrieval by texture similarity.,''
\newblock {\em Pattern Recognition}, vol. 36, no. 3, pp. 665--679, 2003.

\bibitem{146}
T.M. Lehmann, Mark.O. G{\"u}ld, T.~Deselaers, D.~Keysers, H.~Schubert,
  K.~Spitzer, H.~Ney, and B.B. Wein,
\newblock ``Automatic categorization of medical images for content based
  retrieval and data mining,''
\newblock {\em Computerized Medical Imaging and Graphics}, vol. 29, no. 2-3,
  pp. 143--155, 2005.

\bibitem{147}
Y.F. Fathabad and M.~Balafar,
\newblock ``Application of content based image retrieval in diagnosis brain
  disease,''
\newblock {\em Int. Journal on Technical and Physical Problems of Engineering},
  vol. 4, no. 4, pp. 122--128, 2012.

\bibitem{148}
Anil~K. Jain and Farshid Farrokhnia,
\newblock ``Unsupervised texture segmentation using gabor filters,''
\newblock {\em Pattern Recognition}, vol. 24, no. 12, pp. 1167--1186, 1991.

\bibitem{23}
P.~Howarth and S.~R{\"u}ger,
\newblock ``Evaluation of texture features for content-based image retrieval,''
\newblock in {\em International Conference on Image and Video Retrieval,
  Springer-Verlag}, 2004.

\bibitem{ldp-lbp-16}
Timo Ojala, Matti Pietik\"{a}inen, and David Harwood,
\newblock ``{A comparative study of texture measures with classification based
  on featured distributions},''
\newblock {\em Pattern Recognition}, vol. 29, no. 1, pp. 51--59, 1996.

\bibitem{17}
Timo Ojala, Matti Pietik\"{a}inen, and Topi M\"{a}enp\"{a}\"{a},
\newblock ``Multiresolution gray-scale and rotation invariant texture
  classification with local binary patterns,''
\newblock {\em IEEE Trans. Pattern Anal. Mach. Intell.}, vol. 24, no. 7, pp.
  971--987, 2002.

\bibitem{18}
H.~M{\"u}ller, P.~Clough, T.~Deselaers, and B.~Caputo,
\newblock {\em ImageCLEF: Experimental Evaluation in Visual Information
  Retrieval},
\newblock Springer, 2010.

\bibitem{20}
P.~J. Phillips, H.~Moon, P.~Rauss, and S.A. Rizvi,
\newblock ``The feret evaluation methodology for face-recognition algorithms,''
\newblock in {\em Proceedings of the 1997 Conference on Computer Vision and
  Pattern Recognition (CVPR '97)}, 1997, CVPR '97, pp. 137--143.

\bibitem{12}
L.~Nanni and A.~Lumini,
\newblock ``Local binary patterns for a hybrid fingerprint matcher,''
\newblock {\em Pattern Recognition}, vol. 41, no. 11, pp. 3461 -- 3466, 2008b.

\bibitem{2}
L.~Nanni and A.~Luminia,
\newblock ``{A reliable method for cell phenotype image classification},''
\newblock {\em Artificial Intelligence in Medicine}, vol. 43, no. 2, pp.
  87--97, June 2008a.

\bibitem{6}
Devrim Unay and Ahmet Ekin,
\newblock ``Intensity versus texture for medical image search and retrieval.,''
\newblock in {\em ISBI}. 2008, pp. 241--244, IEEE.

\bibitem{Tizhoosh2015}
H.R.Tizhoosh,
\newblock ``Barcode annotations for medical image retrieval,''
\newblock in {\em proceedings of EEE ICIP 2015, Quebec City, Canada}, 2015.

\bibitem{14}
Hichem Frigui, Joshua Caudill, and A.~C.~Ben Abdallah,
\newblock ``Fusion of multi-modal features for efficient content-based image
  retrieval,''
\newblock in {\em IEEE International Conference on Fuzzy Systems}, 2008, pp.
  1992--1998.

\bibitem{59}
C.~G.~M. Snoek, M.~Worring, and A.W.~M. Smeulders,
\newblock ``Early versus late fusion in semantic video analysis,''
\newblock in {\em The 13th Annual ACM International Conference on Multimedia},
  2005, pp. 399--402.

\bibitem{local}
Andrew MacFarlane and Andrew Tuson,
\newblock ``Local search: A guide for the information retrieval practitioner,''
\newblock {\em Information Processing and Management}, vol. 45, no. 1, pp. 159
  -- 174, 2009.

\bibitem{unay1}
C.J.C. Burges,
\newblock ``A tutorial on support vector machines for pattern recognition,''
\newblock {\em Data Min. Knowl. Discov.}, vol. 2(2), pp. 121--167, 1998.

\bibitem{Mueller2010}
H.~M{\"u}ller, P.~Clough, T.~Deselaers, and B.~Caputo,
\newblock {\em Overview of the CLEF 2009 Medical Image Annotation Track},
\newblock Springer, NY, USA, 2010.

\bibitem{CT}
J.~Umamaheswari and G.~Radhamani,
\newblock ``Quadratic program optimization using support vector machine for ct
  brain image classification,'' 2012.

\bibitem{Bishop}
C.M. Bishop,
\newblock {\em Pattern Recognition and Machine Learning (Information Science
  and Statistics)},
\newblock Springer-Verlag New York, Inc., NJ, USA, 2006.

\bibitem{boost}
H.J. Xing, J.G. Wu, and X.F. Chen,
\newblock ``Modified adaboost based ocsvm ensemble for image retrieval,''
\newblock in {\em International Conference on Machine Learning and
  Cybernetics}, 2012, vol.~3, pp. 1048--1053.

\bibitem{Rumelhart1986}
D.E. Rumelhart, G.E. Hinton, , and R.J. Williams,
\newblock ``Learning internal representations by error propagation,''
\newblock {\em Parallel Distributed Processing}, vol. 1, 1986.

\bibitem{Vincent2008}
P.~Vincent, H.~Larochelle, Y.~Bengio, and P.A. Manzagol,
\newblock ``Extracting and composing robust features with denoising
  autoencoders,''
\newblock in {\em International Conference on Machine Learning}, 2008, pp.
  1096--1103.

\bibitem{Krizhevsky2011}
Alex Krizhevsky and Geoffrey~E. Hinton,
\newblock ``Using very deep autoencoders for content-based image retrieval.,''
\newblock in {\em ESANN}, 2011.

\bibitem{Tan2011}
C.~C. Tan and Eswaran C.,
\newblock ``Using autoencoders for mammogram compression,''
\newblock {\em J. Med. Syst.}, vol. 35(1), pp. 49--58, 2011.

\bibitem{Valentine2012}
A.P. Valentine and J.~Trampert,
\newblock ``Data space reduction, quality assessment and searching of
  seismograms: autoencoder networks for waveform data,''
\newblock {\em Geophysical J. International}, vol. 189 (2), pp. 1183--1202,
  2012.

\bibitem{Bengio2013}
Y.~Bengio, A.~Courville, and P.~Vincent,
\newblock ``Representation learning: A review and new perspectives,''
\newblock {\em IEEE Transactions on Pattern Analysis and Machine Intelligence},
  vol. 35, no. 8, pp. 1798--1828, 2013.

\bibitem{Wang2014}
W.~Wang, Y.~Huang, Y.~Wang, and L.~Wang,
\newblock ``Generalized autoencoder: A neural network framework for
  dimensionality reduction,''
\newblock in {\em IEEE Conf. on Computer Vision and Pattern Recognition}, 2014,
  pp. 496--503.

\bibitem{Baldi2012}
P.~Baldi,
\newblock ``Autoencoders, unsupervised learning, and deep architectures,''
\newblock {\em JMLR: Workshop and Conference Proceedings 27}, pp. 37--50, 2012.

\bibitem{Lu2014}
Y.~Lu, L.~Zhang, B.~Wang, and J.~Yang,
\newblock ``Feature ensemble learning based on sparse autoencoders for image
  classification,''
\newblock in {\em International Joint Conference on Neural Networks}, 2014, pp.
  1739--1745.

\bibitem{Tizhoosh2007a}
M.~Ventresca and H.R. Tizhoosh,
\newblock ``Simulated annealing with opposite neighbors,''
\newblock in {\em Foundations of Computational Intelligence, 2007. FOCI 2007.
  IEEE Symposium on}, 2007, pp. 186--192.

\bibitem{Tizhoosh2007b}
M.~Ventresca and H.R. Tizhoosh,
\newblock ``Opposite transfer functions and backpropagation through time,''
\newblock in {\em IEEE Symposium on Foundations of Computational Intelligence
  FOCI 2007}, 2007, pp. 570--577.

\bibitem{Tizhoosh2008}
S.~Rahnamayan and H.R. Tizhoosh,
\newblock ``Image thresholding using micro opposition-based differential
  evolution (micro-ode),''
\newblock in {\em Evolutionary Computation, 2008. CEC 2008. (IEEE World
  Congress on Computational Intelligence). IEEE Congress on}, 2008, pp.
  1409--1416.

\bibitem{libsvm}
C.C. Chang and C.J. Lin,
\newblock ``{LIBSVM}: A library for support vector machines,''
\newblock {\em ACM Transactions on Intelligent Systems and Technology}, vol. 2,
  pp. 27--27, 2011.

\end{thebibliography}
\end{document}